\documentclass[11pt]{article}
\usepackage[preprint]{acl}
\usepackage{times,latexsym}
\usepackage[T1]{fontenc}
\usepackage{amsmath,amssymb,graphicx,booktabs,tabularx}
\usepackage{placeins}
\title{FractureVLM: An Interactive System for Physics-Grounded Multimodal Reasoning and Visual World Generation}

\author{Qinwu Xu, Xiaofu Ma, Yifan Jiang}

\begin{document}
\maketitle
\begin{abstract}
Scientific multimodal systems must connect visual observations and natural language to physical representations without asserting unsupported mechanisms. We present the design of FractureVLM, an interactive system for physics-grounded visual reasoning and exploratory visual world generation. The system combines reusable textual physical rules and visual demonstrations with applicability-aware structured reasoning, supporting scientific explanations and material-conditioned fracture visualization. Crystallographic Miller indices provide interpretable geometric constraints when lattice information is sufficient; otherwise the system abstains from assigning a global index. Qualitative demonstrations cover ceramic and glass fracture, including three paired generation examples and published laboratory reference photographs for ceramic impacts. Observed differences suggest potential benefits of physical conditioning, but also reveal controllability limitations. The interactive implementation and quantitative static evaluation remain pending; the intended use is scientific visualization rather than quantitative mechanical simulation.
\end{abstract}
\section{Introduction}
Multimodal large language models (MLLMs) can interpret images and generate scientific descriptions, but fluent outputs need not be grounded in observable evidence or physical laws. Scientific visualization also requires translating language instructions into visual sequences without conflating plausibility with physical fidelity. Fracture provides a testbed for both directions of multimodal grounding. In crystalline solids, cleavage may follow crystallographic planes described by Miller indices \citep{cullity2001,anderson2017}; in amorphous or heterogeneous materials, a global crystallographic representation may be inappropriate. Even for crystalline materials, the orientation may be unidentifiable from a projected image.

The geometric interpretation of Miller indices provides a bridge from crystallographic physics to numerical fracture modeling. In finite element modeling of carbon-fiber composite machining, \citet{usui2014finite} employed indexed candidate fracture planes within a structured mesh and material-dependent failure criteria, with numerical results evaluated against cutting experiments. This illustrates how indexed plane orientations can be abstracted from crystallographic lattice planes to approximate fracture geometry in a discrete numerical representation. Inspired by this transition, we transfer geometric and material-specific physical knowledge into multimodal prompts for qualitative fracture reasoning and visual generation. For crystalline materials, Miller indices provide additional constraints when lattice information is available; for amorphous glass, which lacks long-range crystalline order, general brittle-fracture principles guide crack initiation, propagation, branching, fragmentation and debris dispersion without assigning crystallographic indices. Thus, the progression from crystallographic physics through finite element modeling to VLM-based visualization motivates a unified, material-aware framework, rather than a universal crystallographic fracture rule. The numerical precedent motivates our representation but does not establish the mechanical accuracy of generated sequences.

Large-scale video pretraining exposes generators to recurring patterns of motion, contact and object interaction. Such statistical regularities may explain plausible outputs even without additional scientific instructions. Physics-IQ shows that existing video generators can reproduce some physical phenomena but that visual realism is not a reliable indicator of physical understanding \citep{motamed2025}. VideoPhy and PhyGenBench similarly document substantial failures in physical commonsense and prompt adherence \citep{bansal2025,meng2025}. On the understanding side, PhysBench identifies limitations in VLM reasoning about object properties and physical dynamics \citep{chow2025}.

A complementary route is to make physical knowledge explicit during generation. VideoDirectorGPT demonstrates that an LLM can expand a prompt into intermediate scene descriptions and spatial layouts that condition a downstream video generator \citep{lin2024}. In diffusion-based systems, such conditioning can steer denoising without itself constituting a mechanical solver. PhysGen combines image-derived physical parameters, rigid-body simulation and diffusion refinement \citep{liu2024physgen}, while WISA incorporates textual, categorical and quantitative physical attributes into video generation \citep{wang2025wisa}. These studies establish precedents for planning and physical conditioning, not evidence that the proprietary generator used here implements a particular internal planning or diffusion architecture.

Our approach complements learned physical regularities with explicit, material-specific scientific instructions and geometric visual examples at inference time, without updating model weights. Fracture provides a particularly challenging setting because damage initiation, crack evolution, fragmentation and subsequent motion must remain mutually consistent. 
We design FractureVLM to bridge visual evidence, natural language and structured physical representations. The intended applications include exploratory scientific visualization and virtual-world content generation, not stress--strain prediction or numerical fracture simulation. The proposed contributions are (i) applicability-aware vision-to-language reasoning with explicit physical hypotheses; (ii) reusable multimodal physical conditioning for language-to-vision generation; and (iii) an interactive workflow with geometric checks and reference-based evaluation. We distinguish completed qualitative generation experiments from the proposed interactive implementation and quantitative reasoning evaluation.
\section{Related Work}
\paragraph{Multimodal reasoning.} Contrastive visual-language learning and visual instruction tuning enable cross-modal understanding \citep{radford2021,liu2023}. Multimodal reasoning systems can produce explanations and interact with visual evidence \citep{yang2023,gpt4,gemini}. Prior work on preference optimization and OCR-aware reasoning motivates grounding generated claims in observations \citep{xu2026,xuetal2026}. Unlike general visual recognition, our task requires deciding whether the requested scientific representation is meaningful and identifiable.
\paragraph{Crystallographic fracture.} Miller indices characterize lattice-plane orientation \citep{cullity2001}. Orientation-dependent cleavage is important in single-crystal fracture, including silicon indentation \citep{skalka2021,wasmer2005}. Directional phase-field formulations likewise model anisotropic fracture \citep{clayton2015,shanthraj2017}. These physical models motivate our structured constraints, but our objective is evaluating generated scientific language, not replacing numerical fracture simulation.
\section{Physical Representation and Task Formulation}
Let $x=(I,T,Q)$ denote a visual observation, available material and coordinate context, and a question. Let $z=(h,k,l)$ denote a candidate crystallographic plane. In a cubic lattice with known axes, a plane can be represented in normalized coordinates by
\begin{equation} hx+ky+lz=d. \end{equation}
Here the coordinate symbols in the equation denote spatial coordinates, and $d$ is the plane offset. For a general lattice, the plane normal is defined through the reciprocal lattice rather than directly by the Cartesian vector $(h,k,l)$.
The intersection with a unit cube yields a polygon dependent on orientation and offset. Projection and viewpoint further affect its appearance. Consequently, polygon shape alone does not establish a unique Miller index.
\paragraph{From crystallographic physics to numerical and visual representations.} Miller indices originate in crystallography, where indexed lattice planes describe orientation-dependent cleavage in crystalline solids, including crystalline constituents of ceramics such as alumina. In polycrystalline ceramics, the orientation may vary among grains. The geometric concept can also be abstracted from the lattice: Usui et al.\ used Miller-index-defined candidate fracture planes in a structured finite element mesh for carbon-fiber composite machining, together with cohesive element insertion and fiber-oriented crack propagation, and compared the numerical procedure with cutting experiments at different fiber orientations \citep{usui2014finite}. This provides an experimentally evaluated precedent for transferring indexed plane geometry to discrete fracture approximations outside crystallographic cleavage. Inspired by that abstraction, we transfer plane geometry, material-specific fracture behavior and loading information into textual and visual conditioning for VLM reasoning and generation. The broader target is qualitative fracture morphology and evolution across brittle materials, including ceramics and amorphous glass. For glass, geometric fracture surfaces and general brittle-fracture rules remain meaningful, but actual crystallographic indices are not assigned. Unlike finite element analysis, our approach does not discretize and solve the mechanical governing equations; the prior numerical validation motivates the representation rather than validating our generated videos.
\paragraph{Applicability and identifiability.} We define four response states $a\in\{G,L,N,U\}$: globally applicable and identifiable ($G$), locally applicable but no justified global plane ($L$), physically inapplicable ($N$), and insufficient evidence ($U$). These are response states, not exhaustive material classes. In particular, crystalline materials may belong to $U$ when lattice orientation is unavailable.
\paragraph{Latent inference.} For identifiable global cases, predict $\hat z=\arg\max_z p(z\mid x,a=G)$. Symmetry-equivalent families should be accepted when the question requests a family rather than an oriented plane.
\paragraph{Consistency.} For a fragment $I_f$ and candidate plane rendering $I_p$, predict compatible, incompatible, or underdetermined. Compatibility does not imply uniqueness.

\section{System Architecture and Implementation}
\begin{figure}[!bthp]
\centering
\setlength{\fboxsep}{4pt}
\begin{tabular}{c}
\fbox{\parbox{.88\columnwidth}{\centering GUI: image, question or scenario}}\\[2pt] $\downarrow$\\[2pt]
\fbox{\parbox{.88\columnwidth}{\centering REST API / request router}}\\[2pt] $\downarrow$\\[2pt]
\fbox{\parbox{.88\columnwidth}{\centering Prompt engine: fixed text rules + visual examples}}\\[2pt] $\downarrow$\\[2pt]
\fbox{\parbox{.88\columnwidth}{\centering VLM: evidence, applicability, physical hypothesis}}\\[2pt] $\downarrow$\\[2pt]
\fbox{\parbox{.88\columnwidth}{\centering Applicability gate / analytical geometry check}}\\[2pt] $\swarrow\qquad\searrow$\\[2pt]
\begin{tabular}{cc}\fbox{\parbox{.38\columnwidth}{\centering Scientific explanation}}&\fbox{\parbox{.38\columnwidth}{\centering Video generator}}\end{tabular}\\[2pt] $\searrow\qquad\swarrow$\\[2pt]
\fbox{\parbox{.88\columnwidth}{\centering Results and comparison interface}}
\end{tabular}
\caption{Proposed system architecture. The fixed physical context is stored in the backend; users supply only task-specific inputs. The video generator may be a separate model.}\label{fig:architecture}
\end{figure}
\paragraph{Interactive workflow.} Figure~\ref{fig:architecture} outlines a proposed Gradio interface and FastAPI backend. Static mode accepts a target image, optional material and coordinate metadata, and a scientific question. Dynamic mode accepts a material, object geometry and loading description. A prompt-mode switch enables baseline and physics-guided comparisons. The backend stores the fixed physical text and three visual demonstrations; users do not upload these repeatedly. Model adapters handle the input format of each reasoning or generation model. Results include structured explanations, applicability, uncertainty and, in dynamic mode, generated video and representative frames. Implementation details, model identifiers, interface screenshot and demo URL remain to be added.
\subsection{Physical and geometric context}
\begin{figure}[!bthp]\centering\includegraphics[width=0.95\linewidth]{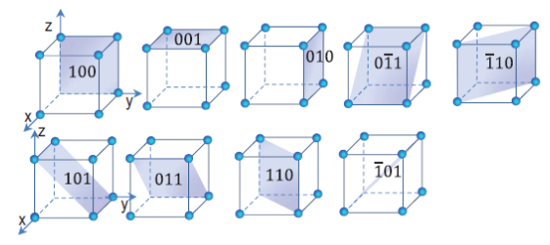}\caption{Representative low-index plane orientations in cubic coordinates. The illustration shows plane configurations, not the complete matched/mismatched evaluation pairs.}\label{fig:pairs}\end{figure}

\begin{figure}[!bthp]\centering\includegraphics[width=1\linewidth]{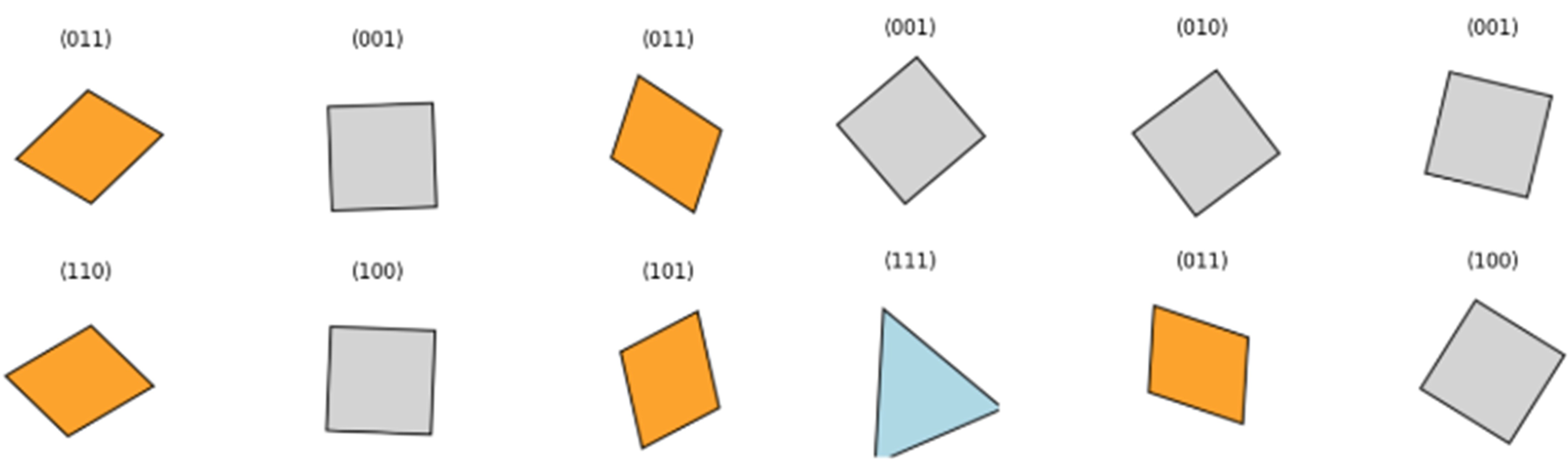}\caption{Labeled synthetic polygon examples from selected cube--plane intersections. Shape alone does not uniquely identify a plane.}\label{fig:fragments}\end{figure}

Figure~\ref{fig:pairs} illustrates representative low-index plane orientations in cubic coordinates. These orientations provide the geometric foundation for the discrete fracture-plane representation discussed above. Figure~\ref{fig:fragments} illustrates selected synthetic cube--plane intersections and their polygonal geometry, providing controlled examples for evaluating geometric reasoning. These examples motivate the visual demonstrations used in the subsequent prompting stage; polygon shape alone does not uniquely determine a Miller index.

\subsection{Physics-guided multimodal in-context prompting}
\begin{figure}[!htbp]
\centering
\includegraphics[width=\linewidth]{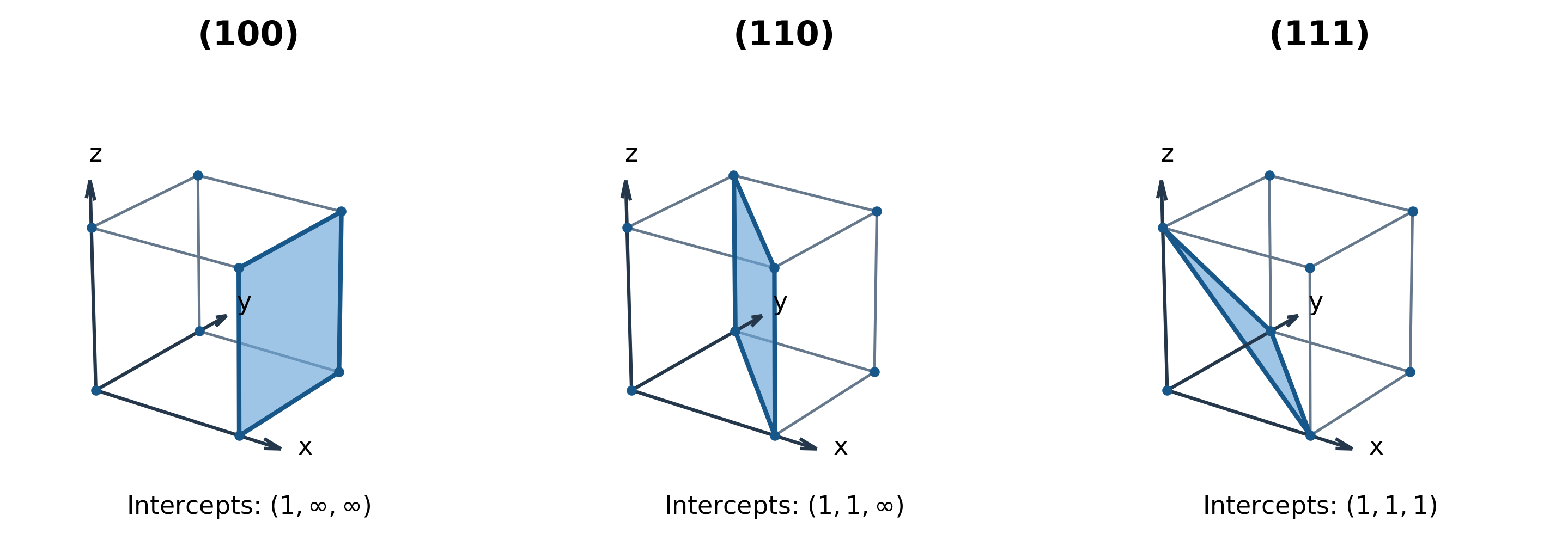}
\caption{Proposed visual in-context demonstrations for canonical cubic planes. The shaded polygons are intersections at $d=1$; axis intercepts illustrate the reciprocal-intercept rule. The reference is supplied to the model alongside the target image.}
\label{fig:prompt_examples}
\end{figure}

We propose supplying a textual physical prior $P$ and labeled visual demonstrations $\mathcal D=\{(I_i,z_i,r_i)\}_{i=1}^{K}$, where $I_i$ is a reference rendering, $z_i$ its index, and $r_i$ its intercept explanation. Figure~\ref{fig:prompt_examples} shows three canonical demonstrations. The reference image is passed as an actual visual input, followed by the unlabeled target image(s) and task instruction. With $C=(x,P,\mathcal D)$, the response distribution becomes $p(a,z,y\mid C)=p(a\mid C)p(z\mid a,C)p(y\mid a,z,C)$. The demonstrations are not a closed candidate set.
\paragraph{Shared prompt template.} The following template combines the original static reasoning instructions with the dynamic generation prompts used in our experiments. Task-specific fields are instantiated for each scenario; the three canonical visual examples are fixed backend context for crystallographic tasks and guided ceramic generation.

\noindent\textit{You are a physics-grounded multimodal reasoning and visualization system. Use the supplied image, material, geometry, loading and boundary conditions, when available. In normalized cubic coordinates, $hx+ky+lz=d$; indices are proportional to reciprocal intercepts, with zero denoting a parallel axis and negative values denoting signed orientations. Use the labeled (100), (110), and (111) diagrams as representative examples, not a closed candidate set.}\par
\noindent\textit{For reasoning, extract visible facets, edges, axes and intercepts. Assess applicability and uncertainty; infer indices only with sufficient evidence, accounting for offset, viewpoint and lattice orientation. Return evidence, applicability, candidate indices or null, compatibility and explanation.}\par
\noindent\textit{For generation, depict [material, specimen geometry, projectile, loading, boundary conditions] from an intact state through impact, crack initiation, propagation, branching and fragmentation. Maintain specimen geometry, impact location and camera viewpoint. Prevent spontaneous crack healing or fragment reconnection. For glass, use qualitative brittle-fracture constraints: initiate damage at contact, propagate cracks outward and separate fragments after cracks connect. Do not impose cubic planes or assign crystallographic indices without appropriate lattice evidence. Do not invent measurements or claim calibrated mechanical accuracy.}

The template summarizes the shared methodology rather than reproducing any one submitted prompt verbatim. Conventional prompts retain the basic scenario and continuity instructions but omit the added case-specific physical context and crystallographic guidance. We plan to release the exact submitted prompts, visual context and available generation logs in the accompanying repository.

\paragraph{Cross-modal reasoning and generation.} Let $C=(I,q,T,P,\mathcal D)$ include an optional image $I$, language instruction $q$, material metadata $T$, physical text $P$ and fixed visual demonstrations $\mathcal D$. We propose the structured reasoning factorization
\begin{equation}
p(a,z,y\mid C)=p(a\mid C)p(z\mid a,C)p(y\mid a,z,C).
\end{equation}
Here $y$ is a grounded language explanation. For dynamic generation, a model adapter may condition a separate generator on the accepted physical representation and language description, $p(V\mid C,a,z,y)$, where $V$ denotes the generated frame sequence. This is a proposed conditioning interface, not a claim of access to internal neural latent activations or of a jointly trained model. When $a=N$ or $U$, no unsupported global index is passed to the generator; local hypotheses for $L$ require grain-level evidence.
\paragraph{Analytical verification.} For controlled cubic geometry, intersect the candidate plane with the twelve cube edges, deduplicate intersections and order the resulting vertices. Under known offset and camera transformation, compare the projected polygon with target geometry. Missing lattice or camera information yields an undetermined result rather than a forced rejection. Material rules reject global crystallographic assignments to known amorphous materials. This proposed independent check is intended to provide functionality beyond unconstrained API prompting; implementation and evaluation remain pending.
\section{System Demonstration and Evaluation}
\subsection{Static visual reasoning}

\begin{figure*}[!t]
\centering
\includegraphics[width=\textwidth]{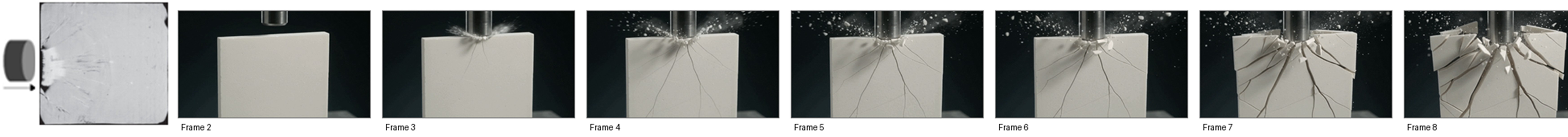}
\caption{Edge-on alumina impact visualization and laboratory reference from \citet{pd2023}. No baseline-versus-guided comparison is claimed.}
\label{fig:case1}
\end{figure*}

\begin{figure}[!t]
\centering
\includegraphics[width=0.49\textwidth]{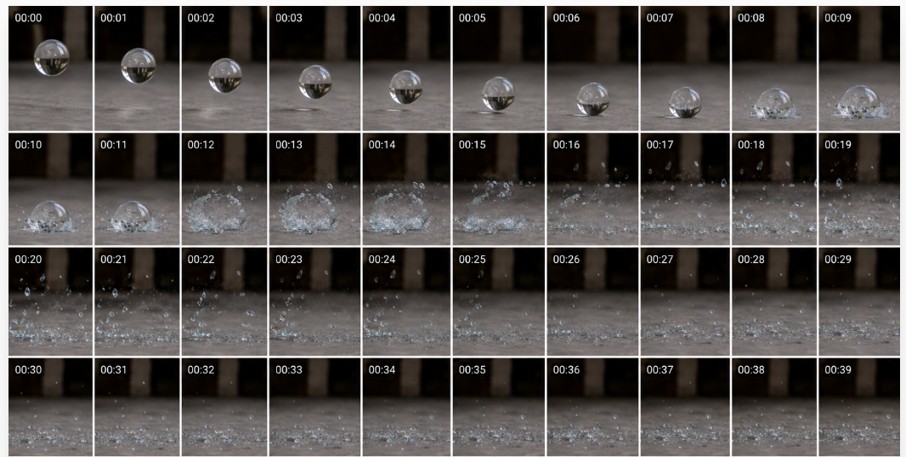}
\caption{Generated fracture progression of a water-filled glass sphere under impact, illustrating approach, fragmentation and debris dispersion. Frame annotations are not calibrated physical time.}
\label{fig:glassball}
\end{figure}

Existing synthetic examples (Figures~\ref{fig:pairs} and~\ref{fig:fragments}) suggest recognition of canonical plane families and selected incompatible pairs. Polygon shape alone is insufficient for unique inference. We plan a paired evaluation on approximately 20--30 held-out cases with known geometry or independently specified applicability. Compare direct prompting, textual priors, combined text and visual priors, and the full verification pipeline using identical target inputs and decoding settings. Report index accuracy on identifiable cases, applicability accuracy, compatibility accuracy, unsupported-claim rate and coverage. Separate invalid claims from underdetermined claims and report false refusals. No aggregate results are available yet.
\subsection{Qualitative Evaluation of Dynamic Fracture Generation}
\label{sec:dynamic_evaluation}
We consider five scenarios grouped into general demonstrations and baseline comparisons. The new runs use the Gemini web interface, which returned 30-frame, 640 $\times$ 360 looping GIFs; eight representative frames are sampled per run. Published laboratory photographs provide external morphology references for ceramic impacts \citep{pd2023,zhang2022}; they are not numerical predictions or time-resolved validation targets.

\paragraph{General demonstrations.} Figures~\ref{fig:case1} and~\ref{fig:glassball} illustrate a cylindrical projectile striking a $70\times70\times10$~mm alumina plate edge-on at 85~m/s \citep{pd2023} and impact-induced fracture of a water-filled glass sphere, respectively. The ceramic example uses the representative conventional run; the guided run changed the impact direction and is not used as a successful edge-impact demonstration. The sphere illustrates fragmentation and debris dispersion. Together, these examples illustrate the broader visualization objective motivated by discrete numerical fracture modeling \citep{usui2014finite}: transferring geometric and material-specific physical knowledge across brittle materials. In the glass example, general brittle-fracture rules, rather than crystallographic Miller-index constraints, provide the physical guidance.

\paragraph{Comparative evaluation.} Cases 1--3 compare conventional and physics-informed prompting. For ceramics, the guided prompts add specimen dimensions, impact speed and available boundary conditions, together with applicability-aware crystallographic context. The bottle uses general brittle-fracture guidance without Miller indices.

\begin{figure*}[!t]
\centering
\includegraphics[width=\textwidth]{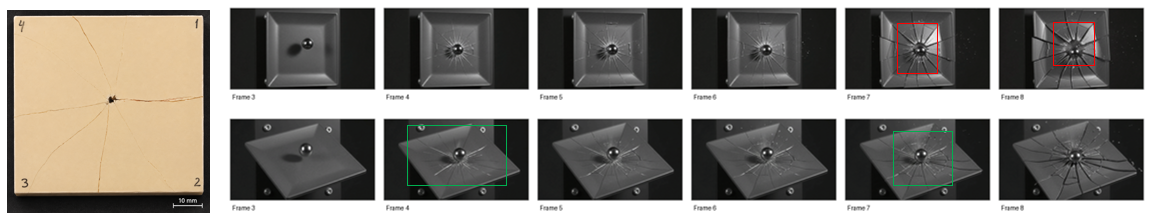}
\caption{Laboratory fracture reference \citep{pd2023} and conventional (top) versus physics-informed (bottom) generation for normal alumina impact at 200~m/s. Both outputs exhibit radial cracks. The conventional generation develops a pronounced circular damage region around the impact site (red boxes), whereas the physics-informed generation exhibits more extended radial cracking and irregular, interconnected fracture morphology (green boxes). Nevertheless, both generations overpredict crack density and late-stage fracture severity relative to the laboratory reference, indicating partial improvement in fracture morphology rather than quantitative mechanical agreement.}
\label{fig:case2}
\end{figure*}

\textbf{Case 1: Alumina normal impact.} A 6.4~mm steel sphere strikes a $70\times70\times9$~mm plate at 200~m/s \citep{pd2023}. Both outputs exhibit central damage and radial cracking, broadly consistent with the laboratory morphology (Figure~\ref{fig:case2}). The guided sequence also depicts localized fragmentation and interconnected cracks; differences in viewpoint and apparent crack opening preclude a quantitative comparison.

\begin{figure*}[!t]
\centering
\includegraphics[width=\textwidth]{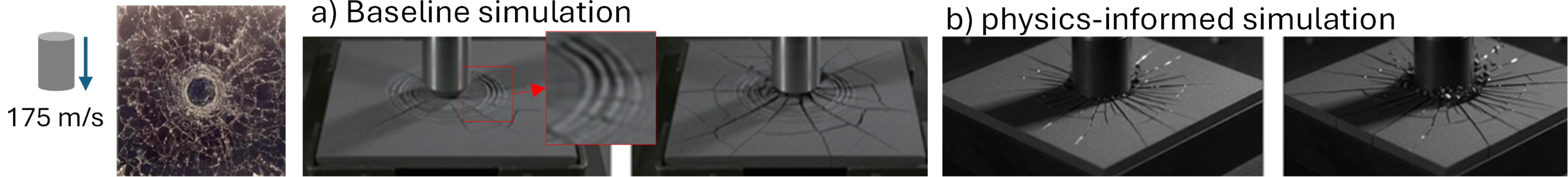}
\caption{Case 2: Laboratory fracture reference \citep{zhang2022}
and comparison of baseline and physics-guided generation for 
silicon carbide plate impact at 175~m/s. The baseline generates 
pronounced wave-like concentric rings, as highlighted in the 
enlarged region, which are less consistent with the experimental 
brittle-fracture morphology. In contrast, the physics-guided 
generation exhibits more distinct radial cracking and localized 
fragmentation, showing better qualitative agreement with the 
experimental fracture pattern.}
\label{fig:case3}
\end{figure*}

\textbf{Case 2: Silicon carbide impact.} A 10~mm diameter, 15~mm high steel cylinder strikes a $60\times60\times8$~mm plate at 175~m/s \citep{zhang2022}. The baseline depicts pronounced wave-like rings, whereas the guided result displays more distinct radial cracking and localized damage (Figure~\ref{fig:case3}). The latter qualitatively resembles aspects of the laboratory fracture morphology; the rings alone do not establish a physical violation.

\begin{figure*}[!bthp]
\centering
\includegraphics[width=\textwidth]{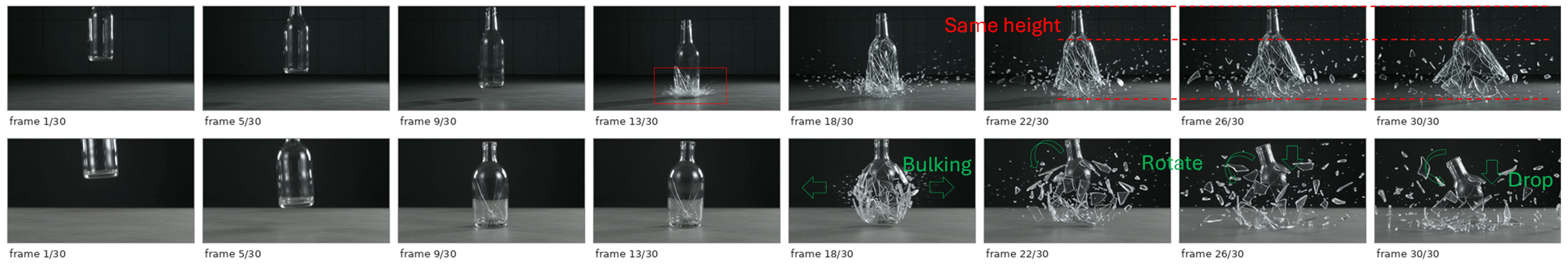}
\caption{Case 3: Comparison of baseline (upper row) and 
physics-guided (lower row) generation of a glass bottle falling 
vertically onto concrete. Eight frames are sampled from each 
30-frame sequence. The baseline exhibits crystal-like 
fragmentation artifacts at frame 13/30 and little apparent 
downward displacement from frames 22/30 to 30/30, with 
fragmentation predominantly concentrated near the bottom. 
In contrast, the physics-guided generation depicts progressive 
fracture, lateral fragment dispersion from the middle region, 
and subsequent downward motion of the remaining upper portion 
under gravity, accompanied by rotation consistent with 
asymmetric structural damage and loss of support. These 
observations suggest improved fracture morphology and temporal 
physical consistency. Frame indices represent generation 
progression rather than calibrated physical time.}
\label{fig:case4}
\end{figure*}

\textbf{Case 3: Glass bottle impact.} The baseline shows crystal-like fragmentation artifacts at frame 13/30 and little apparent downward displacement during frames 22/30--30/30 despite damage concentrated near the base. The guided sequence depicts lateral fragmentation from the middle region, followed by downward motion of the remaining upper portion and rotation consistent with asymmetric damage and loss of support (Figure~\ref{fig:case4}). These observations suggest greater temporal plausibility in this example, although the sequence does not reach a settled fragment field. No matched laboratory reference is available for this case.

The paired examples are exploratory: one run per prompt, differing camera viewpoints and uncontrolled sampling prevent causal attribution. The ceramic references do not establish lattice orientation, so these experiments assess general physical plausibility rather than direct Miller-index control.

\section{Discussion and Conclusion}
FractureVLM is designed to connect natural language and visual evidence through explicit physical representations, supporting both grounded explanations and exploratory visual world generation. Its central hypothesis is that selective physical conditioning and independent verification improve interpretability and reduce unsupported scientific claims. The ceramic reference cases offer an external comparison for visual morphology, while synthetic crystalline geometry provides analytically verifiable reasoning targets. The paired qualitative results suggest potential improvements in morphology and temporal coherence, while showing failures of geometric controllability. They do not establish a controlled prompt effect. Completing the interactive implementation and quantitative static evaluation remains necessary before submission.
\section*{Limitations}
The system is not a numerical fracture solver. Unknown lattice orientation prevents global index validation in the ceramic reference cases. Generated videos cannot establish conservation laws or calibrated mechanics. 


\FloatBarrier

\end{document}